\documentclass[11pt]{article}

\usepackage[preprint]{acl}

\usepackage{times}
\usepackage{latexsym}
\usepackage{booktabs}      % \toprule, \midrule, \bottomrule, \cmidrule
\usepackage{multirow}      % column spanning in tables (\multicolumn already built-in)
\usepackage{tabularx}      % optional, for auto-width columns if needed
\usepackage{xspace} % for \xspace command
\usepackage[table]{xcolor} % for \rowcolor in tables
\usepackage{tikz}
\usetikzlibrary{positioning, arrows.meta, shapes, fit}

\usepackage[T1]{fontenc}
\usepackage[utf8]{inputenc}

\usepackage{microtype}

\usepackage{inconsolata}

\usepackage{graphicx}

\usepackage{amsmath}
\usepackage{amsfonts}
\usepackage{amssymb}
\usepackage{algorithm}
\usepackage{algorithmic}
\usepackage{tcolorbox}
\usepackage{xcolor}

\usepackage{pifont}
\usepackage{fontawesome5}

\newcommand{\std}[2]{$#1_{\scriptscriptstyle \textcolor{red!70!black}{\pm #2}}$}
\newcommand{\stdb}[2]{$\mathbf{#1}_{\scriptscriptstyle \textcolor{red!70!black}{\pm #2}}$}

\newcommand{\logo}{\raisebox{-0.2em}{\includegraphics[height=1.5em]{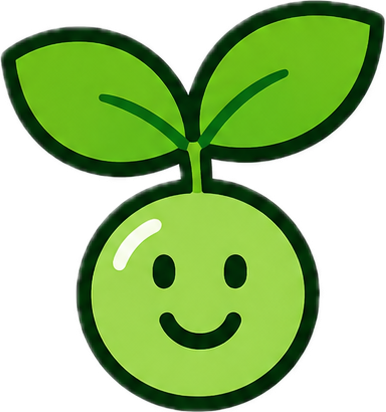}}}
\newcommand{\name}[1]{\textsc{SproutRAG}\xspace}

\title{\logo \name{}: Attention-Guided Tree Search with Progressive Embeddings for Long-Document RAG}

\author{
\textbf{
Amirhossein Abaskohi$^{1}$\thanks{Corresponding author: \texttt{aabaskoh@cs.ubc.ca}},
Issam H. Laradji$^{1,2},$
Peter West$^{1}$,
Giuseppe Carenini$^{1}$}\\[0.5em]
$^{1}$University of British Columbia, $^{2}$ServiceNow Research 
}

\begin{document}
\maketitle

% -----------------------------------------------
% SECTIONS
% -----------------------------------------------

\begin{abstract}
    Retrieval-augmented generation (RAG) systems must balance retrieval granularity with contextual coherence, a challenge that existing methods address through LLM-guided chunking, single-level context expansion, or hierarchical summarization. These approaches variously depend on costly LLM calls during indexing or retrieval, limit context aggregation to a single granularity level, or introduce information loss through summarization. We present \textbf{\name{}}, an attention-guided hierarchical RAG framework that addresses this trade-off by organizing sentence-level chunks into progressively larger but semantically coherent units, using learned inter-sentence attention to construct a binary chunking tree. Unlike prior approaches that rely on external LLMs, fixed context expansion, or lossy summarization, \name{} learns which attention heads and layers best capture semantic document structure, enabling multi-granularity retrieval without additional LLM calls or compressed summaries. At retrieval time, \name{} uses hierarchical beam search to retrieve candidates at multiple granularities, capturing multi-sentence relevance beyond flat retrieval. The framework is trained end-to-end with a joint objective that improves both embeddings and tree structure. Experiments across four benchmarks spanning scientific, legal, and open-domain settings demonstrate that \name{} improves information efficiency (IE) by 6.1\% on average over the strongest baseline\footnote{Code is available on \href{https://github.com/AmirAbaskohi/SproutRAG}
    {{\textcolor{black}{\faGithub}}~GitHub}.}.
\end{abstract}
\section{Introduction}
\label{sec:introduction}

Retrieval-augmented generation (RAG) has become the dominant paradigm for grounding large language models (LLMs) in external knowledge, helping reduce hallucinations, support domain-specific reasoning, and improve performance on knowledge-intensive tasks~\cite{lewis2021rag, augenstein2023factuality}. As LLMs are increasingly applied to complex tasks involving long documents~\cite{jin2025longcontext}, directly providing entire documents as input becomes impractical due to context-length constraints and degraded attention over extended sequences~\cite{jin2024longcontext, liu2023lost}. Consequently, RAG frameworks segment documents into chunks and retrieve the most relevant pieces to construct focused, high-quality evidence for generation.

The effectiveness of this retrieval step hinges critically on how documents are segmented. Large chunks preserve contextual coherence but introduce redundant noise that dilutes key information, while fine-grained chunks offer precision but suffer from semantic fragmentation and broken inter-chunk relationships~\cite{tao2025sakirag, zhao-etal-2025-moc}. This problem is particularly acute for cross-paragraph retrieval, where answering a query requires synthesizing information scattered across multiple document sections, as in multi-hop reasoning and summarization tasks~\cite{liu-etal-2025-hoprag}.

Recent work addresses this challenge from several directions. SAKI-RAG~\cite{tao2025sakirag} uses a SLLM~\cite{an2024sentencevae} to merge semantically related sentence pairs and relies on an external LLM to filter retrieval candidates. However, extending this pairwise expansion to multi-chunk relevance greatly increases the candidate space and makes LLM filtering expensive. LLM-guided chunking methods such as Meta-Chunking~\cite{zhao2024metachunking} and MoC~\cite{zhao-etal-2025-moc} improve segmentation quality, but discard cross-chunk dependencies after chunking. Hierarchical methods such as RAPTOR~\cite{sarthi2024raptor} support multi-granularity retrieval through clustering and summarization, yet clustering treats chunks within a group as interchangeable and summaries can lose evidence. Graph-based methods such as GraphRAG~\cite{edge2024graphrag} model entity relations, but are less effective when fine-grained chunks contain sparse entity information.

\begin{figure}[t]
    \centering
    \includegraphics[width=\linewidth]{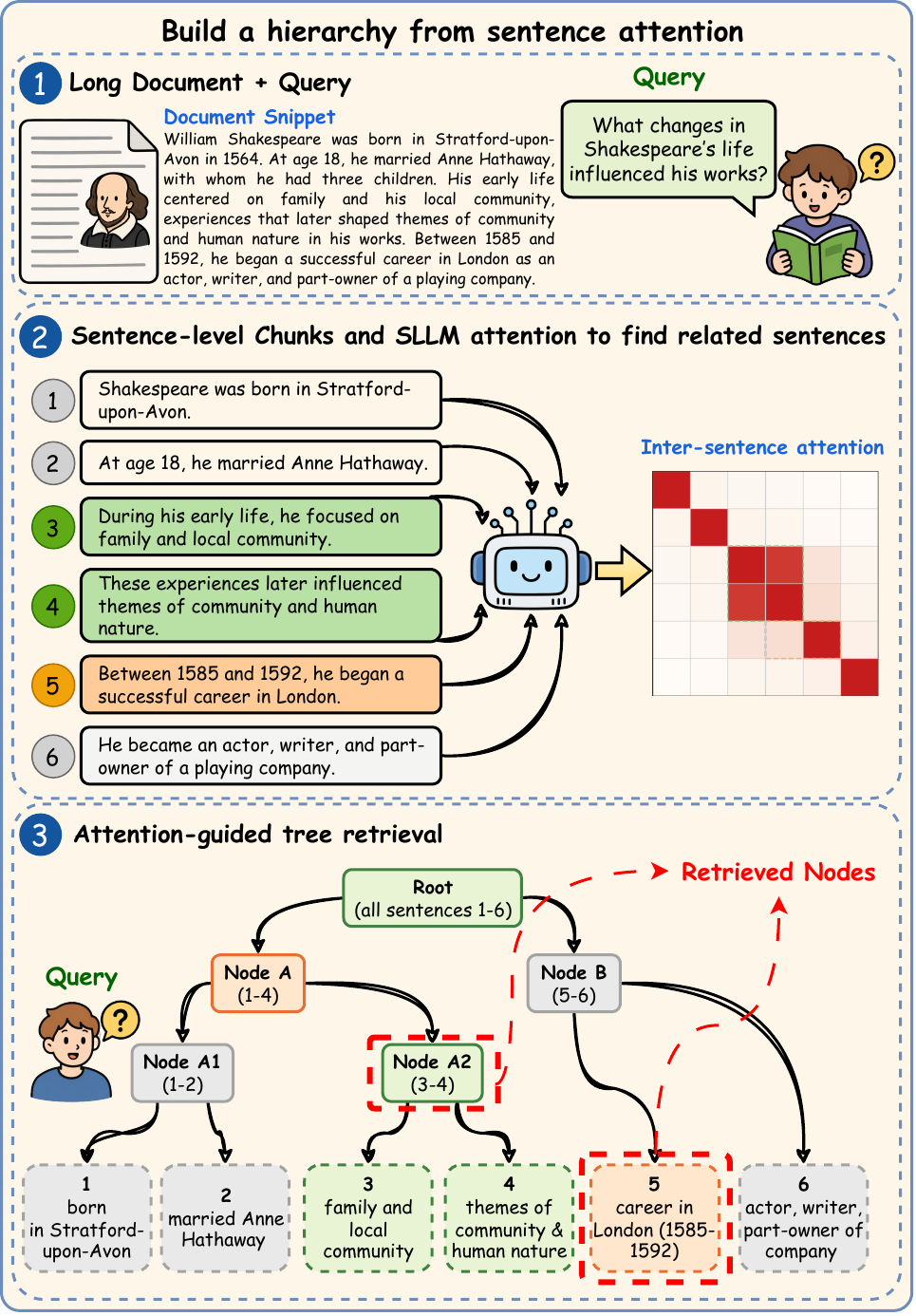}
    \caption{\name{} segments a long document into sentence-level chunks, uses SLLM attention to identify semantically related sentences, and organizes them into an attention-guided binary tree. Retrieval then selects evidence across fine-grained leaves, mid-level nodes, and broader subtrees, preserving precision while recovering coherence.}
    \label{fig:teaser}
\end{figure}

In this paper, we present \textbf{\name{}}, an attention-guided hierarchical RAG framework that organizes sentence-level chunks into a learned document structure, preserving cross-chunk dependencies while avoiding LLM inference overhead during retrieval. As illustrated in Figure~\ref{fig:teaser}, this structure enables retrieval at multiple semantic granularities. \name{} encodes documents at sentence granularity using an SLLM and constructs a binary tree bottom-up, where merge order is determined by a learned weighted aggregation of inter-sentence attention across transformer heads and layers. This aggregation replaces naive uniform averaging, which we show introduces a proximity bias that weakens the global tree index; instead, learnable scalar weights allow the model to discover which head types best reflect semantic co-relevance for document structure. Each internal node stores a progressive embedding that compositionally represents its subtree, enabling multi-granularity retrieval via hierarchical beam search that collects candidates across all tree levels. \name{} is trained end-to-end to jointly optimize retrieval quality and tree structure, requiring no external LLM calls at any stage. It captures emergent multi-sentence relevance that pairwise or flat retrieval methods cannot, while remaining efficient enough for deployment. We evaluate \name{} on four benchmarks spanning scientific, legal, and open-domain settings. On average, \name{} improves information efficiency (IE) by 6.1\% over the strongest baseline, especially in the cases where evidence is often dispersed across paragraphs. 

In summary, the contributions of this paper are as follows. \textbf{(1)} We introduce \name{}, an attention-guided hierarchical RAG framework that constructs a binary tree over sentence-level chunks using learned inter-sentence attention, enabling multi-granularity retrieval without any LLM calls at inference time. \textbf{(2)} We identify and address the proximity bias introduced by uniform attention averaging in sentence-level transformers, replacing it with a learned weighted aggregation that allows the model to discover which attention heads best reflect semantic co-relevance for document structure. \textbf{(3)} We introduce a joint training objective that jointly improves retrieval quality and tree structure, eliminating the need for external LLM filtering or lossy summarization at any pipeline stage.
\section{Related Work}
\label{sec:related_work}

\paragraph{Chunking and adaptive retrieval.}
The effectiveness of RAG depends strongly on how documents are segmented into retrievable units. Standard RAG pipelines often rely on rule-based splitters, such as fixed-length or delimiter-based chunking, which are efficient but insensitive to semantic boundaries~\cite{team2024langchain}. Recent methods aim to improve this granularity choice. Late-Chunking~\cite{gunther2024latechunking} contextualizes token representations before forming chunk embeddings, while Meta-Chunking~\cite{zhao2024metachunking} and MoC~\cite{zhao-etal-2025-moc} use LLM-based signals or routing mechanisms to produce more adaptive chunk boundaries. Dense X Retrieval~\cite{chen2024densex} moves toward finer granularity by decomposing text into atomic propositions, improving precision but weakening broader contextual continuity. Other methods adapt retrieval after chunks have been formed. ReflectiveRAG~\cite{verma-etal-2026-reflectiverag} introduces a self-reflective retrieval loop that evaluates evidence sufficiency and reformulates queries to improve factual grounding, but it does not change the underlying flat organization of retrieval units. Most related to \name{}, SAKI-RAG~\cite{tao2025sakirag} uses a SLLM to estimate inter-sentence attention and expand retrieved chunks with related sentences. Unlike SAKI-RAG's pairwise expansion with LLM filtering, \name{} builds a global sentence-level hierarchy that supports multi-granularity retrieval without inference-time LLM calls.

\paragraph{Structured and hierarchical retrieval.}
Beyond flat chunk retrieval, structured RAG methods organize document content into higher-level representations. RAPTOR~\cite{sarthi2024raptor} recursively clusters chunks and summarizes each cluster into a tree, enabling retrieval from multiple levels; however, its structure is based on embedding-space clustering and relies on LLM-generated summaries, which can discard fine-grained details. Graph-based approaches such as GraphRAG~\cite{edge2024graphrag} and LightRAG~\cite{guo2025lightrag} represent documents through entities and relations, supporting traversal-based retrieval but depending on successful entity extraction and relation construction. PropRAG~\cite{wang2025proprag} replaces entity triples with propositions and performs LLM-free beam search over proposition paths, while Beam Retrieval~\cite{zhang-etal-2024-end} shows the benefit of maintaining multiple retrieval hypotheses for multi-hop passage retrieval. PageIndex~\cite{zhang2025pageindex} similarly explores reasoning-based, vectorless retrieval over document tree structures, but relies on document-level structural organization rather than learned sentence-level attention. \name{} instead builds an attention-guided binary tree over sentence-level chunks, with compositional internal nodes and joint retrieval over all nodes. This preserves cross-sentence dependencies without lossy summarization, entity-centric structures, or external LLM calls.
\section{\name{}}
\label{sec:method}

\begin{figure*}[t]
    \centering
    \includegraphics[width=\textwidth]{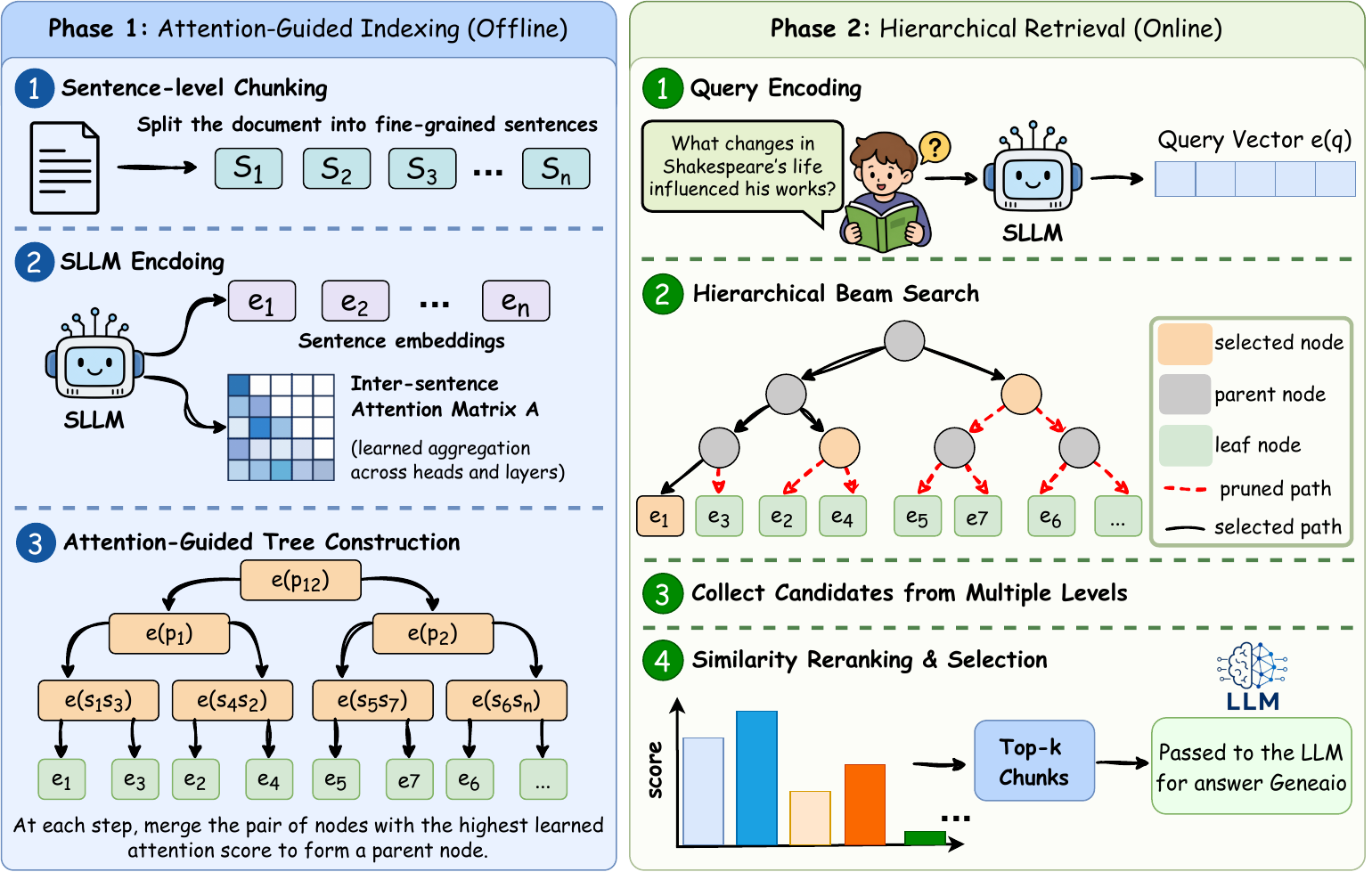}
    \caption{
    \textbf{Overview of \name{}.} In the offline indexing phase (\textbf{Phase 1}), documents are split into sentence-level chunks and encoded with a SLLM to obtain sentence embeddings and inter-sentence attention. Learned aggregation over attention heads and layers guides bottom-up tree construction, producing an attention tree with sentence embeddings at the leaves and progressive embeddings at internal nodes. In the online retrieval phase(\textbf{Phase 2}), a query is encoded and hierarchical beam search traverses the tree, collecting candidates from multiple levels before similarity reranking selects the top-$k$ chunks for answer generation.
    }
    \label{fig:sprout-overview}
\end{figure*}

As illustrated in Figure~\ref{fig:sprout-overview}, \name{} replaces flat chunk retrieval with a \textbf{trained attention-guided hierarchy} over sentence-level chunks. During offline indexing, a SLLM encodes the document and provides both sentence embeddings and inter-sentence attention signals. These signals are aggregated with learnable head--layer weights and used to build a binary tree, where leaves represent fine-grained chunks and internal nodes store \textbf{progressive embeddings} of merged sentence groups. During online retrieval, \name{} encodes the query and performs \textbf{hierarchical beam search}, collecting candidates from leaves, internal nodes, and subtrees. As described in Section~\ref{sec:training}, the framework is trained with a joint objective that improves both \textbf{retrieval quality} and \textbf{tree structure}, enabling multi-granularity retrieval without external LLM calls during retrieval.

\subsection{Attention-Guided Indexing}
\label{sec:indexing}

Given a document $D$, we first split it into sentence-level chunks
$S=\{s_1,\ldots,s_n\}$. We encode the full sequence with a SLLM, obtaining contextualized sentence embeddings $\{e(s_i)\}_{i=1}^{n}$ and attention matrices from all layers and heads. For layer $l$ and head $h$, we denote the corresponding attention matrix as $\mathrm{Attn}^{(l,h)} \in \mathbb{R}^{n \times n}$.

A uniform average over all heads and layers can overemphasize local sentence proximity, since some attention heads primarily capture sequential patterns~\cite{voita-etal-2019-analyzing}. To reduce this \textbf{proximity bias}, \name{} learns a weighted aggregation over heads and layers:
\begin{equation}
    \mathbf{A}_{ij}
    =
    \sum_{l=1}^{L}
    \sum_{h=1}^{H}
    w_{l,h}\,\mathrm{Attn}^{(l,h)}_{ij},
\label{eq:attention_aggregation}
\end{equation}
where $w_{l,h}$ is defined as:
\begin{equation}
    w_{l,h}
    =
    \frac{\exp(\alpha_{l,h})}
    {\sum_{l'=1}^{L}\sum_{h'=1}^{H} \exp(\alpha_{l',h'})}.
\label{eq:attention_weights}
\end{equation}
The learnable scalars $\alpha_{l,h}$ allow the model to emphasize attention heads that better capture semantic co-relevance. We then symmetrize the aggregated attention to obtain a mutual relation score:
\begin{equation}
    \mathbf{M}_{ij}
    =
    \frac{\mathbf{A}_{ij}+\mathbf{A}_{ji}}{2}.
    \label{eq:mutual_attention}
\end{equation}

The tree is built bottom-up. Initially, each sentence chunk is a leaf node. At each step, we merge the pair of active nodes with the highest mutual attention score. The parent embedding is computed as a \textbf{progressive embedding} of its children:
\begin{equation}
    e(p)
    =
    \frac{e(u)+e(v)}{2},
    \label{eq:progressive_embedding}
\end{equation}
where $u$ and $v$ are the merged child nodes. After merging, the new parent inherits its strongest relation to each remaining node:
\begin{equation}
    \mathbf{M}_{pr}
    =
    \max(\mathbf{M}_{ur}, \mathbf{M}_{vr}).
    \label{eq:linkage}
\end{equation}
This single-linkage update preserves long-range semantic connections as the hierarchy grows. The result is an \textbf{attention tree} $\mathcal{T}$ whose leaves retain sentence-level precision and whose internal nodes represent broader semantic units.

\subsection{Hierarchical Retrieval}
\label{sec:retrieval}

Given a query $q$, \name{} first encodes it with the same SLLM used during indexing to obtain a query embedding $e(q)$. Retrieval is then performed over the attention-guided binary tree, where each node $v$ represents a document span at a specific granularity. Leaf nodes correspond to sentence-level chunks, while internal nodes represent progressively larger groups of semantically related sentences. This allows \name{} to retrieve evidence at the level most appropriate for the query, rather than relying on a fixed chunk size.

Each candidate node is scored by cosine similarity between the query embedding and the node representation:
\begin{equation}
    \mathrm{sim}(q,v)
    =
    \frac{e(q)^\top e(v)}
    {\|e(q)\|\,\|e(v)\|}.
    \label{eq:similarity}
\end{equation}

Starting from the root node, retrieval proceeds via hierarchical beam search. Let $\mathcal{B}_t$ denote the active beam at depth $t$, with $\mathcal{B}_0=\{v_{\mathrm{root}}\}$. At each step, \name{} expands the children of the current beam nodes and retains the top-$b$ most relevant nodes:
\begin{equation}
    \mathcal{B}_{t+1}
    =
    \operatorname{Top}_b
    \left(
        \bigcup_{v \in \mathcal{B}_t} \mathrm{Child}(v),
        \mathrm{sim}(q,\cdot)
    \right),
    \label{eq:beam_update}
\end{equation}
where $b$ is the beam width. This search strategy focuses computation on the most promising branches of the tree while still allowing the retriever to explore multiple semantically relevant regions of the document.

In parallel, \name{} collects relevant nodes encountered during traversal. Let $\mathcal{V}_{\mathrm{visit}}$ denote the set of all nodes scored during beam search:
\begin{equation}
    \mathcal{V}_{\mathrm{visit}}
    =
    \bigcup_t
    \bigcup_{v \in \mathcal{B}_t}
    \mathrm{Child}(v).
\end{equation}
The retrieval candidate set is then defined as all visited nodes whose similarity exceeds a \textbf{learned threshold $\delta$}:
\begin{equation}
    \mathcal{C}
    =
    \left\{
        v \in \mathcal{V}_{\mathrm{visit}}
        \;:\;
        \mathrm{sim}(q,v) \geq \delta
    \right\}.
    \label{eq:candidate_collection}
\end{equation}

The candidate set $\mathcal{C}$ contains evidence at multiple granularities, from sentence-level leaves to larger subtrees, allowing \name{} to retrieve either precise facts or broader multi-sentence context as needed. The collected candidates are reranked by similarity or a lightweight reranker, and the top-$k$ chunks are passed to the answer generator. The complete indexing and retrieval procedure is summarized in Algorithm~\ref{alg:sproutrag}.

\begin{algorithm}[t]
\caption{\name{} Indexing and Retrieval}
\label{alg:sproutrag}
\begin{algorithmic}[1]
\REQUIRE Document $D$, query $q$, beam width $b$, threshold $\delta$, top-$k$
\ENSURE Retrieved evidence $F$

\STATE \COMMENT{\textbf{Offline indexing}}
\STATE $S \leftarrow \mathrm{SplitSentences}(D)$
\STATE $E, \mathcal{A} \leftarrow \mathrm{SLLM}(S)$
\STATE $\mathbf{A} \leftarrow \sum_{l,h} w_{l,h}\,\mathcal{A}^{(l,h)}$
\STATE $\mathbf{M} \leftarrow (\mathbf{A}+\mathbf{A}^{\top})/2$

\STATE $\mathcal{N} \leftarrow \{ \mathrm{Leaf}(s_i, e_i) \mid s_i \in S, e_i \in E \}$
\WHILE{$|\mathcal{N}| > 1$}
    \STATE $(u,v) \leftarrow \arg\max_{u \neq v;\, u,v \in \mathcal{N}} \mathbf{M}_{uv}$
    \STATE $p \leftarrow \mathrm{Node}(u,v)$
    \STATE $e_p \leftarrow (e_u + e_v)/2$
    \FOR{$r \in \mathcal{N} \setminus \{u,v\}$}
        \STATE $\mathbf{M}_{pr} \leftarrow \max(\mathbf{M}_{ur}, \mathbf{M}_{vr})$
        \STATE $\mathbf{M}_{rp} \leftarrow \mathbf{M}_{pr}$
    \ENDFOR
    \STATE $\mathcal{N} \leftarrow (\mathcal{N} \setminus \{u,v\}) \cup \{p\}$
\ENDWHILE
\STATE $\mathcal{T} \leftarrow \mathrm{root}(\mathcal{N})$

\STATE \COMMENT{\textbf{Online retrieval}}
\STATE $e_q \leftarrow \mathrm{SLLM}(q)$
\STATE $\mathcal{C} \leftarrow \emptyset,\quad \mathcal{B} \leftarrow \{\mathcal{T}\}$
\WHILE{$\mathcal{B} \neq \emptyset$}
    \STATE $\mathcal{C} \leftarrow \mathcal{C} \cup \{v \in \mathcal{B} \mid \mathrm{sim}(e_q,e_v) \geq \delta\}$
    \STATE $\mathcal{X} \leftarrow \bigcup_{v \in \mathcal{B}} \mathrm{Children}(v)$
    \STATE $\mathcal{B} \leftarrow \mathrm{TopB}(\mathcal{X}, b, \mathrm{sim}(e_q,\cdot))$
\ENDWHILE

\STATE $F \leftarrow \mathrm{TopK}(\mathrm{Rerank}(\mathcal{C}, q), k)$
\RETURN $F$
\end{algorithmic}
\end{algorithm}

\subsection{Joint Training}
\label{sec:training}

The pretrained SLLM is not optimized for retrieval or for constructing retrieval-oriented document structures. We therefore fine-tune \name{} with a joint objective that improves both the embedding space and the attention tree.

\paragraph{Retrieval objective.}
We train the SLLM embeddings with contrastive learning over query--passage pairs. Given a query $q$, a positive passage $p^+$, and hard negatives $\{p_j\}$, each passage is represented by mean-pooling its sentence embeddings. We optimize:
\begin{equation}
    \mathcal{L}_{\mathrm{ret}}
    =
    -\log
    \frac{
        \exp(\mathrm{sim}(q,p^+)/\tau)
    }{
        \sum_j \exp(\mathrm{sim}(q,p_j)/\tau)
    },
    \label{eq:retrieval_loss}
\end{equation}
where $\tau$ is a temperature parameter. This objective aligns queries with relevant passages and separates them from hard negatives.

\paragraph{Structure objective.}
Good embeddings alone do not guarantee a useful hierarchy. Since the tree depends on the learned attention matrix, we add an attention regularizer that encourages co-relevant sentence pairs to receive high mutual attention. Let $\mathcal{G}$ be the set of sentence pairs within a positive passage that jointly support the query. We define:
\begin{equation}
    \mathcal{L}_{\mathrm{attn}}
    =
    -
    \frac{1}{|\mathcal{G}|}
    \sum_{(s_i,s_j)\in\mathcal{G}}
    \log
    \left(
    \frac{\mathbf{A}_{ij}+\mathbf{A}_{ji}}{2}
    \right).
    \label{eq:attention_loss}
\end{equation}
This objective directly shapes the learned head--layer aggregation, encouraging the induced tree to group semantically related evidence into coherent and retrievable subtrees.

\paragraph{Final objective.}
The final training loss is:
\begin{equation}
    \mathcal{L}
    =
    \mathcal{L}_{\mathrm{ret}}
    +
    \lambda \mathcal{L}_{\mathrm{attn}},
    \label{eq:joint_loss}
\end{equation}
where $\lambda$ controls the strength of structure regularization. After training, the learned aggregation weights are used during offline indexing, and retrieval requires only query encoding, tree traversal, and reranking. Thus, \name{} avoids external LLM filtering and lossy LLM-based summarization while enabling efficient multi-granularity retrieval.
\section{Experiments and Results}

\subsection{Experimental Setup}
\label{sec:experimental_setup}

\noindent\textbf{Benchmarks.}
We evaluate \name{} on four retrieval benchmarks spanning scientific, legal, and open-domain settings: SCI-DOCS~\cite{cohan-etal-2020-specter}, LegalBench-RAG~\cite{pipitone2024legalbenchragbenchmarkretrievalaugmentedgeneration}, Dragonball~\cite{zhu-etal-2025-rageval}, and MS MARCO~\cite{bajaj2018msmarco}. For end-to-end answer generation, we further evaluate on HotpotQA~\cite{yang-etal-2018-hotpotqa}, WebQuestions~\cite{berant-etal-2013-semantic}, and Dragonball. See Appendix~\ref{app:benchmark_details} for more details.

\noindent\textbf{Baselines.}
We compare against representative chunking and structured retrieval methods, including Dense X Retrieval~\cite{chen2024densex}, Meta-Chunking~\cite{zhao2024metachunking}, MoC~\cite{zhao-etal-2025-moc}, RAPTOR~\cite{sarthi2024raptor}, LightRAG~\cite{guo2025lightrag}, PropRAG~\cite{wang2025proprag}, and SAKI-RAG~\cite{tao2025sakirag}. GraphRAG~\cite{edge2024graphrag}, ReflectiveRAG~\cite{verma-etal-2026-reflectiverag}, PageIndex~\cite{zhang2025pageindex}, and REFRAG~\cite{lin2025refrag} are reported only for final task performance, as they primarily involve LLM-heavy reasoning, generation, summarization, or decoding-time optimization rather than efficient retrieval. For fair comparison, methods requiring an LLM or reranker use the same \textsc{Qwen3-8B}~\cite{qwen3technicalreport} generator and \textsc{Qwen3-Reranker-4B}~\cite{zhang2025qwen3embeddingadvancingtext} reranker. Other than unifying the generator and reranker, we follow the settings recommended in the original papers for each baseline. See Appendix~\ref{app:baseline_details} for more details.

\noindent\textbf{Metrics.}
For retrieval evaluation, we report Recall, Precision, and Information Efficiency (IE), where $\mathrm{IE}=\mathrm{Recall}\times\mathrm{Precision}$. To keep the evidence budget comparable across methods, $k$ denotes the number of underlying evidence units used for evaluation rather than the number of retrieved tree nodes. For hierarchical outputs, retrieved internal nodes are expanded into their underlying evidence units, with each contained unit counted toward the same top-$k$ budget. We compute each metric at $k\in\{1,3,5\}$ and report the average across these three cutoffs. For end-to-end generation, we report F1 on HotpotQA and WebQuestions, and ROUGE-L~\cite{lin-2004-rouge}, METEOR~\cite{banerjee-lavie-2005-meteor}, and BERTScore~\cite{Zhang2020BERTScore} on Dragonball. We also report online efficiency using Tok/Q and latency, where Tok/Q counts online model-token usage per query, excluding offline indexing and output tokens. All reported results are averaged over three independent runs.

\noindent\textbf{Implementation and Training Details.}
We use the 1.3B-parameter SLLM~\cite{an2024sentencevae}\footnote{\url{https://github.com/cavedweller509/SentenceVAE}} as the sentence encoder and split documents into chunks of up to two sentences. \name{} is trained on 30K query--passage examples sampled from CLaRa~\cite{he2025clarabridgingretrievalgeneration}, fine-tuning the SLLM and learned head--layer aggregation weights with the joint objective in Eq.~\ref{eq:joint_loss}. Unless otherwise stated, we use the following hyperparameters as the default setting: 3 training epochs with AdamW, learning rates of $2\times10^{-5}$ for the SLLM and $1\times10^{-3}$ for the aggregation scalars, batch size 32, temperature $\tau=0.05$, attention weight $\lambda=0.1$, and 5\% linear warmup. At retrieval time, the default setting uses beam width $b=5$, collects candidates from all tree levels, and reranks them with \textsc{Qwen3-Reranker-4B}. Final answers are generated with \textsc{Qwen3-8B}. For fair comparison, baselines that require an LLM or reranker use the same models; otherwise, we follow their original settings. All experiments use 8 NVIDIA A100 80GB GPUs, and results are averaged over three runs.

\begin{table*}[t]
    \centering
    \resizebox{\textwidth}{!}{%
    \begin{tabular}{lccc ccc}
    \toprule
    \multirow{2}{*}{\textbf{Method}}
    & \multicolumn{3}{c}{\textbf{Dragonball}}
    & \multicolumn{3}{c}{\textbf{SCI-DOCS}} \\
    \cmidrule(lr){2-4}
    \cmidrule(lr){5-7}
    & \textbf{Rec.} $\uparrow$ & \textbf{Pre.} $\uparrow$ & \textbf{IE} $\uparrow$
    & \textbf{Rec.} $\uparrow$ & \textbf{Pre.} $\uparrow$ & \textbf{IE} $\uparrow$ \\
    \midrule
    Dense X Retrieval
    & \std{3.18}{0.09} & \std{5.76}{0.14} & \std{0.18}{0.02}
    & \std{95.64}{0.31} & \std{88.13}{0.42} & \std{84.31}{0.47} \\
    
    Meta-Chunking-PPL
    & \std{42.91}{0.54} & \std{44.62}{0.63} & \std{19.14}{0.37}
    & \std{22.16}{0.71} & \std{18.73}{0.64} & \std{4.15}{0.19} \\
    
    Meta-Chunking-MSP
    & \std{32.74}{0.47} & \std{42.08}{0.58} & \std{13.77}{0.31}
    & \std{96.37}{0.29} & \std{88.42}{0.46} & \std{85.21}{0.43} \\
    
    RAPTOR
    & \std{32.19}{0.61} & \std{41.92}{0.76} & \std{13.49}{0.34}
    & \std{97.72}{0.24} & \std{93.03}{0.39} & \std{90.92}{0.41} \\
    
    LightRAG
    & \std{47.38}{0.69} & \std{54.16}{0.72} & \std{25.67}{0.46}
    & \std{97.51}{0.27} & \std{92.84}{0.44} & \std{90.53}{0.45} \\
    
    PropRAG
    & \std{49.81}{0.64} & \std{58.24}{0.79} & \std{29.01}{0.52}
    & \std{98.07}{0.22} & \std{94.36}{0.35} & \std{92.55}{0.38} \\
    
    MoC
    & \std{51.26}{0.57} & \std{60.47}{0.68} & \std{30.99}{0.44}
    & \std{98.18}{0.19} & \std{94.73}{0.31} & \std{93.01}{0.34} \\
    
    SAKI-RAG
    & \std{36.83}{0.52} & \std{78.61}{0.64} & \std{28.95}{0.41}
    & \std{91.82}{0.33} & \std{97.43}{0.28} & \std{89.46}{0.36} \\
    \midrule
    \rowcolor{blue!8}
    \textbf{\name{}}
    & \stdb{45.76}{0.43} & \stdb{85.34}{0.51} & \stdb{39.05}{0.38}
    & \stdb{98.74}{0.16} & \stdb{98.91}{0.18} & \stdb{97.66}{0.21} \\
    \bottomrule
    \end{tabular}%
    }

    \vspace{2pt}
    
    \resizebox{\textwidth}{!}{%
    \begin{tabular}{lccc ccc}
    \toprule
    \multirow{2}{*}{\textbf{Method}}
    & \multicolumn{3}{c}{\textbf{LegalBench-RAG}}
    & \multicolumn{3}{c}{\textbf{MS MARCO}} \\
    \cmidrule(lr){2-4}
    \cmidrule(lr){5-7}
    & \textbf{Rec.} $\uparrow$ & \textbf{Pre.} $\uparrow$ & \textbf{IE} $\uparrow$
    & \textbf{Rec.} $\uparrow$ & \textbf{Pre.} $\uparrow$ & \textbf{IE} $\uparrow$ \\
    \midrule
    Dense X Retrieval
    & \std{24.83}{0.38} & \std{35.74}{0.51} & \std{8.88}{0.21}
    & \std{58.41}{0.46} & \std{62.68}{0.55} & \std{36.61}{0.48} \\
    
    Meta-Chunking-PPL
    & \std{27.82}{0.44} & \std{38.41}{0.57} & \std{10.69}{0.26}
    & \std{60.84}{0.52} & \std{63.97}{0.49} & \std{38.92}{0.45} \\
    
    Meta-Chunking-MSP
    & \std{28.31}{0.41} & \std{39.26}{0.49} & \std{11.12}{0.24}
    & \std{61.73}{0.43} & \std{64.36}{0.52} & \std{39.73}{0.39} \\
    
    RAPTOR
    & \std{28.64}{0.48} & \std{39.83}{0.62} & \std{11.41}{0.29}
    & \std{62.48}{0.56} & \std{65.81}{0.67} & \std{41.12}{0.53} \\
    
    LightRAG
    & \std{30.62}{0.53} & \std{43.97}{0.58} & \std{13.47}{0.33}
    & \std{66.13}{0.51} & \std{68.42}{0.61} & \std{45.25}{0.57} \\
    
    PropRAG
    & \std{31.47}{0.46} & \std{45.28}{0.54} & \std{14.25}{0.32}
    & \std{66.92}{0.48} & \std{69.83}{0.58} & \std{46.73}{0.49} \\
    
    MoC
    & \std{32.18}{0.42} & \std{46.11}{0.51} & \std{14.84}{0.27}
    & \std{67.31}{0.45} & \std{70.26}{0.53} & \std{47.29}{0.46} \\
    
    SAKI-RAG
    & \std{31.38}{0.39} & \std{46.27}{0.47} & \std{14.52}{0.25}
    & \std{68.04}{0.42} & \std{71.58}{0.49} & \std{48.70}{0.44} \\
    \midrule
    \rowcolor{blue!8}
    \textbf{\name{}}
    & \stdb{36.91}{0.34} & \stdb{53.48}{0.41} & \stdb{19.74}{0.29}
    & \stdb{72.86}{0.37} & \stdb{76.21}{0.44} & \stdb{55.53}{0.40} \\
    \bottomrule
    \end{tabular}%
    }
    
    \caption{
    Retrieval performance across four benchmarks. Recall, Precision, and IE are averaged over @1, @3, and @5, with IE computed at each cutoff before averaging. Values report the mean over three independent runs, and the red $\textcolor{red!70!black}{\pm}$ values indicate the corresponding standard deviation. The shaded row marks \name{}. Refer to Appendix~\ref{app:detailed_results} for the results @1, @3, and @5.
    }
    \label{tab:retrieval_results}
\end{table*}

\begin{table*}[t]
    \centering
    \resizebox{\textwidth}{!}{%
    \begin{tabular}{l
      c c
      c c c
      c c}
    \toprule
    \multirow{3}{*}{\textbf{Method}}
    & \multicolumn{5}{c}{\textbf{Final Performance}}
    & \multicolumn{2}{c}{} \\
    \cmidrule(lr){2-6}
    & \textbf{HotpotQA}
    & \textbf{WebQuestions}
    & \multicolumn{3}{c}{\textbf{Dragonball}}
    & \multicolumn{2}{c}{\textbf{Average Cost}} \\
    \cmidrule(lr){2-2}
    \cmidrule(lr){3-3}
    \cmidrule(lr){4-6}
    \cmidrule(lr){7-8}
    & \textbf{F1} $\uparrow$
    & \textbf{F1} $\uparrow$
    & \textbf{R-L} $\uparrow$
    & \textbf{MTR} $\uparrow$
    & \textbf{BRT} $\uparrow$
    & \textbf{Tok/Q} $\downarrow$
    & \textbf{Lat. (ms)} $\downarrow$ \\
    \midrule
    GraphRAG
    & 72.18 & 64.73
    & 0.346 & 0.361 & 0.637
    & 16238 & 2317 \\
    
    ReflectiveRAG
    & 70.64 & 63.29
    & 0.334 & 0.349 & 0.622
    & 11274 & 1186 \\
    
    REFRAG
    & 73.42 & 65.38
    & 0.351 & 0.368 & 0.641
    & 5436 & 492 \\
    
    PageIndex
    & \textbf{79.36} & \textbf{70.81}
    & \textbf{0.389} & \textbf{0.406} & \textbf{0.691}
    & 24620 & 2847 \\
    \midrule
    \rowcolor{blue!8}
    \textbf{\name{}}
    & 76.47 & 68.12
    & 0.372 & 0.389 & 0.671
    & \textbf{4382} & \textbf{193} \\
    \bottomrule
    \end{tabular}%
    }
    \caption{
    End-to-end answer quality and online efficiency. HotpotQA and WebQuestions are evaluated with F1, while Dragonball uses ROUGE-L (R-L), METEOR (MTR), and BERTScore (BRT). Tok/Q counts online model input tokens per query, excluding offline training, indexing, and output tokens; Lat. reports online per-query latency. All methods use the same generator and reranker when applicable.
    }
    \label{tab:end_to_end_results}
\end{table*}

\subsection{Retrieval Quality}
\label{sec:retrieval_results}

Table~\ref{tab:retrieval_results} reports retrieval performance across four benchmarks. \name{} achieves the highest IE on all datasets, improving over the strongest baseline by \textbf{8.06} points on Dragonball, \textbf{4.65} on SCI-DOCS, \textbf{4.90} on LegalBench-RAG, and \textbf{6.83} on MS MARCO. These improvements are not driven by recall alone: \name{} also obtains the \textbf{best precision on every benchmark}. This suggests that the attention-guided hierarchy helps retrieve broader supporting context while avoiding the noise introduced by overly large or weakly related chunks.

The comparison with SAKI-RAG is especially informative. While SAKI-RAG achieves strong precision, particularly on Dragonball and SCI-DOCS, its pairwise expansion limits evidence aggregation, reducing recall and IE. In contrast, \name{} converts sentence-level attention into a \textbf{global tree structure}, enabling retrieval over individual chunks, internal nodes, and subtrees. This preserves SAKI-RAG's precision benefits while improving IE across all datasets. Structured baselines such as RAPTOR, LightRAG, PropRAG, and MoC improve recall over flat or boundary-based chunking, but their clustering, graph, proposition, or routing structures do not explicitly model learned multi-sentence composition. \name{} bridges this gap: leaves retain fine-grained evidence, while internal nodes recover coherent context, yielding the strongest recall--precision tradeoff. Appendix~\ref{app:qualitative_legal} provides a qualitative example.

\subsection{End-to-End Performance and Efficiency}

We next evaluate whether the retrieval improvements translate into stronger final answers. Table~\ref{tab:end_to_end_results} compares \name{} with system-level RAG methods on HotpotQA, WebQuestions, and Dragonball. PageIndex achieves the highest final answer scores, but it requires substantially more online computation due to its reasoning-based search and evidence construction. REFRAG improves efficiency compared with reflection-heavy or reasoning-heavy systems, but \name{} still provides the strongest \textbf{performance--efficiency tradeoff}: it outperforms GraphRAG, ReflectiveRAG, and REFRAG across all final-performance metrics, while using only \textbf{4.38K online tokens per query} and \textbf{193 ms} latency. The reported cost measures online per-query inference and excludes offline training and indexing. \textbf{\name{} does require an upfront training stage}: we fine-tune the SLLM and attention aggregation weights on a 30K-example subset of CLaRa. However, \textbf{this cost is paid once and reused across datasets}, similar to other systems with offline preparation or model adaptation costs. The cross-dataset results show that the learned attention-guided hierarchy generalizes without retraining for each benchmark, making the training cost amortized rather than query-time overhead.

\begin{table*}[t]
    \centering
    \resizebox{\textwidth}{!}{%
    \begin{tabular}{lccc ccc ccc ccc}
    \toprule
    \multirow{2}{*}{\textbf{Variant}}
    & \multicolumn{3}{c}{\textbf{Dragonball}}
    & \multicolumn{3}{c}{\textbf{SCI-DOCS}}
    & \multicolumn{3}{c}{\textbf{LegalBench-RAG}}
    & \multicolumn{3}{c}{\textbf{MS MARCO}} \\
    \cmidrule(lr){2-4}
    \cmidrule(lr){5-7}
    \cmidrule(lr){8-10}
    \cmidrule(lr){11-13}
    & \textbf{Rec.} $\uparrow$ & \textbf{Pre.} $\uparrow$ & \textbf{IE} $\uparrow$
    & \textbf{Rec.} $\uparrow$ & \textbf{Pre.} $\uparrow$ & \textbf{IE} $\uparrow$
    & \textbf{Rec.} $\uparrow$ & \textbf{Pre.} $\uparrow$ & \textbf{IE} $\uparrow$
    & \textbf{Rec.} $\uparrow$ & \textbf{Pre.} $\uparrow$ & \textbf{IE} $\uparrow$ \\
    \midrule

    \rowcolor{blue!8}
    \textbf{\name{} ($b=5,\lambda=0.1$)}
    & \textbf{45.76} & \textbf{85.34} & \textbf{39.05}
    & \textbf{98.74} & \textbf{98.91} & \textbf{97.66}
    & \textbf{36.91} & \textbf{53.48} & \textbf{19.74}
    & \textbf{72.86} & \textbf{76.21} & \textbf{55.53} \\

    \midrule
    \rowcolor{gray!15}
    \multicolumn{13}{c}{\textit{\textbf{Training Objectives}}} \\

    \rowcolor{gray!4}
    Not trained
    & 34.28 & 69.41 & 23.79
    & 90.36 & 92.18 & 83.28
    & 29.47 & 42.53 & 12.53
    & 64.18 & 67.42 & 43.28 \\

    \rowcolor{gray!4}
    w/o $\mathcal{L}_{\mathrm{ret}}$
    & 37.64 & 73.52 & 27.67
    & 92.71 & 94.86 & 87.96
    & 31.08 & 45.17 & 14.04
    & 66.24 & 69.73 & 46.18 \\

    \rowcolor{gray!4}
    w/o $\mathcal{L}_{\mathrm{attn}}$
    & 41.39 & 78.26 & 32.39
    & 96.42 & 96.88 & 93.41
    & 33.52 & 48.31 & 16.20
    & 69.18 & 72.64 & 50.25 \\

    \midrule
    \rowcolor{orange!12}
    \multicolumn{13}{c}{\textit{\textbf{Tree and Retrieval Design}}} \\

    \rowcolor{orange!4}
    Uniform attention aggregation
    & 39.18 & 76.94 & 30.15
    & 95.83 & 96.12 & 92.11
    & 32.74 & 47.36 & 15.51
    & 68.41 & 71.52 & 48.92 \\

    \rowcolor{orange!4}
    Embedding-similarity tree
    & 40.72 & 79.38 & 32.32
    & 96.31 & 96.74 & 93.17
    & 33.18 & 48.42 & 16.06
    & 69.36 & 72.48 & 50.28 \\

    \rowcolor{orange!4}
    Leaf-only retrieval
    & 38.26 & 83.19 & 31.83
    & 94.87 & 98.12 & 93.07
    & 31.94 & 52.61 & 16.80
    & 67.83 & 75.42 & 51.16 \\

    \rowcolor{orange!4}
    Greedy search
    & 39.84 & 82.47 & 32.86
    & 96.25 & 98.34 & 94.65
    & 32.81 & 51.76 & 16.98
    & 68.92 & 74.86 & 51.60 \\

    \midrule
    \rowcolor{green!12}
    \multicolumn{13}{c}{\textit{\textbf{Hyperparameter Sensitivity}}} \\

    \rowcolor{green!4}
    Beam width $b=3$
    & 44.12 & 84.71 & 37.37
    & 98.31 & 98.83 & 97.16
    & 35.72 & 52.86 & 18.88
    & 71.94 & 75.63 & 54.42 \\

    \rowcolor{green!4}
    Beam width $b=10$
    & 46.24 & 82.63 & 38.21
    & 98.91 & 98.42 & 97.35
    & 37.28 & 51.92 & 19.36
    & 73.41 & 74.32 & 54.56 \\

    \rowcolor{green!4}
    $\lambda=0.05$
    & 44.68 & 83.92 & 37.50
    & 98.42 & 98.63 & 97.07
    & 35.96 & 52.47 & 18.87
    & 72.14 & 75.48 & 54.46 \\

    \rowcolor{green!4}
    $\lambda=0.20$
    & 45.18 & 82.74 & 37.38
    & 98.58 & 98.37 & 96.97
    & 36.42 & 51.83 & 18.88
    & 72.49 & 74.93 & 54.31 \\
    \bottomrule
    \end{tabular}%
    }
    \caption{
    Ablation study on retrieval performance. Metrics are averaged over @1, @3, and @5. The blue row is the default \name{} setting ($b=5,\lambda=0.1$). The three groups evaluate training objectives, tree/retrieval design, and sensitivity to $b$ and $\lambda$.
    }
    \label{tab:ablation_retrieval}
    \vspace{-1em}
\end{table*}

\subsection{Ablation Study}

\noindent
\textbf{Training objectives. }
In the \colorbox{gray!15}{first group} of Table~\ref{tab:ablation_retrieval},  evaluates the role of the training objectives. The \textit{Not trained} variant performs worst across all datasets, showing that the pretrained SLLM attention and embeddings are not sufficient for retrieval-oriented tree construction. Removing $\mathcal{L}_{\mathrm{ret}}$ substantially reduces both recall and IE, since the query and evidence embeddings are no longer explicitly aligned. Removing $\mathcal{L}_{\mathrm{attn}}$ is less damaging than removing the retrieval loss, but still causes a consistent drop, especially in IE. This confirms that \textbf{embedding quality and tree quality require complementary supervision}: the retrieval loss aligns query--passage representations, while the attention-structure loss shapes the hierarchy used for multi-granularity retrieval.

\noindent
\textbf{Tree and retrieval design. }
The \colorbox{orange!12}{second group} in Table~\ref{tab:ablation_retrieval} examines the necessity of the attention-guided hierarchy. Uniform attention aggregation reduces performance, highlighting that averaging heads and layers introduces \textbf{proximity bias} and weakens the tree. An embedding-similarity tree also underperforms, showing that SLLM attention encodes structural information beyond embeddings. Leaf-only retrieval maintains high precision but lowers recall and IE, while greedy search suffers from early path commitment. These results demonstrate that \textbf{learned attention aggregation}, \textbf{internal-node retrieval}, and \textbf{beam search} are all crucial for balancing precise evidence with broader contextual coverage.

\noindent
\textbf{Hyperparameter sensitivity. }
The \colorbox{green!12}{final group} in Table~\ref{tab:ablation_retrieval} studies beam width $b$ and attention regularization weight $\lambda$. Reducing the beam width to $b=3$ slightly lowers IE because fewer semantic paths are explored, while increasing it to $b=10$ improves recall but slightly reduces precision, yielding no overall advantage over the default $b=5$. Similarly, both $\lambda=0.05$ and $\lambda=0.20$ underperform the default $\lambda=0.1$: a weaker structure loss provides insufficient guidance for tree construction, while a stronger one can overemphasize attention alignment at the expense of retrieval precision. Overall, \name{} is stable across reasonable settings, with $b=5$ and $\lambda=0.1$ providing the best recall--precision tradeoff.
\section{Conclusion and Future Work}
\label{sec:conclusion}

We introduced \name{}, an attention-guided hierarchical RAG framework that organizes sentence-level chunks into a learned tree for multi-granularity retrieval. Rather than relying on fixed chunk boundaries, pairwise context expansion, lossy summarization, or inference-time LLM filtering, \name{} uses learned SLLM attention aggregation to construct a retrieval-oriented hierarchy. At inference time, hierarchical beam search selects evidence from sentence leaves, internal nodes, and broader subtrees, allowing the retriever to balance fine-grained precision with contextual coherence. Across benchmarks, \name{} improves retrieval information efficiency by 6.1\% on average, offering a strong \textbf{performance--efficiency tradeoff} that approaches LLM-heavy systems while using far fewer online tokens and lower latency. While \name{} generalizes well after one-time training, several directions remain open. Future work can explore richer node composition functions beyond mean pooling, such as gated or attention-based composition, and dynamic tree adaptation or query-dependent traversal policies for complex multi-hop retrieval.

% -----------------------------------------------
% Mandatory Sections
% -----------------------------------------------

\section*{Limitations}

While \name{} improves multi-granularity retrieval without inference-time LLM filtering, it has some limitations. First, the hierarchy is currently built as a \textbf{binary tree}, which may be restrictive when several sentences jointly form a coherent semantic unit and should be grouped together simultaneously. Multi-branch trees could better capture such many-to-many dependencies. Second, \name{} requires an upfront training stage for the SLLM and attention aggregation weights. Although this is a one-time cost that transfers across datasets in our experiments, it is still more expensive than using an off-the-shelf retriever without adaptation. Finally, tree construction is offline and fixed during retrieval. While this makes inference efficient and avoids rebuilding the index per query, it may be less flexible when queries require evidence reorganized by query-specific relevance.

% -----------------------------------------------
% References
% -----------------------------------------------

\bibliography{custom}

% -----------------------------------------------
% Appendix
% -----------------------------------------------

\appendix

\section{Benchmark and Baseline Details}
\label{app:benchmark_baseline_details}

\subsection{Benchmark Dataset Details}
\label{app:benchmark_details}

We include four retrieval-focused benchmarks---SCI-DOCS, LegalBench-RAG, Dragonball, and MS MARCO---and three end-to-end generation benchmarks---HotpotQA, WebQuestions, and Dragonball. Together, these datasets cover scientific retrieval, legal retrieval, open-domain passage retrieval, multi-hop question answering, short-form factual QA, and multi-domain RAG evaluation.

\paragraph{SCI-DOCS~\cite{cohan-etal-2020-specter}} is a scientific document representation benchmark introduced with SPECTER. It contains multiple document-level tasks, including citation prediction, document classification, and recommendation. We use SCI-DOCS as a scientific retrieval benchmark because scientific abstracts are dense, terminology-heavy, and often contain multiple related concepts within a short span. This makes the dataset useful for testing whether \name{} can construct coherent sentence-level hierarchies in technical domains.

\paragraph{LegalBench-RAG~\cite{pipitone2024legalbenchragbenchmarkretrievalaugmentedgeneration}} is designed specifically to evaluate retrieval in legal RAG pipelines. It contains 6,858 query-answer pairs over legal documents such as NDAs, M\&A agreements, commercial contracts, and privacy policies. Unlike broad document retrieval, LegalBench-RAG emphasizes precise snippet retrieval: the model must identify minimal legal evidence rather than simply retrieve a generally relevant document. This makes it a strong test of fine-grained precision.

\paragraph{Dragonball~\cite{zhu-etal-2025-rageval}} is a multi-domain and multilingual RAG benchmark released as part of RAGEval. It contains questions across finance, legal, and medical scenarios in English and Chinese. We use Dragonball for both retrieval and end-to-end generation because it combines heterogeneous domains, long evidence contexts, and domain-specific terminology. This setting tests whether retrieval methods can recover relevant evidence without introducing excessive distractor context.

\paragraph{MS MARCO~\cite{bajaj2018msmarco}} is a large-scale open-domain passage retrieval benchmark built from real web search queries. Its corpus consists of millions of short passages, and the task requires identifying passages that answer natural language questions. Compared with SCI-DOCS and LegalBench-RAG, MS MARCO has shorter retrieval units and more direct query-passage matching, providing a complementary test of retrieval effectiveness when evidence is already compact.

\paragraph{HotpotQA~\cite{yang-etal-2018-hotpotqa}} is an open-domain multi-hop QA benchmark built from Wikipedia. Its questions require reasoning over multiple supporting documents or facts, and the dataset provides sentence-level supporting-fact annotations. We use HotpotQA for end-to-end answer generation because it directly tests whether retrieved evidence supports multi-step reasoning, which aligns with \name{}'s goal of retrieving evidence across multiple granularities.

\paragraph{WebQuestions~\cite{berant-etal-2013-semantic}} is an open-domain factual QA benchmark built from natural language questions collected from web search logs. The answers are typically short entities or phrases, making token-level F1 a suitable evaluation metric. We include WebQuestions to evaluate whether \name{} also improves short-form factual QA, where retrieval must remain precise and avoid adding unnecessary context.

\subsection{Baseline Details}
\label{app:baseline_details}

We compare \name{} against two groups of baselines: efficient retrieval-oriented methods used in the retrieval evaluation, and system-level RAG methods used in the end-to-end generation comparison. Unless otherwise stated, we follow the configurations recommended in the original papers. For methods requiring an LLM generator or reranker, we use the same \textsc{Qwen3-8B} generator and \textsc{Qwen3-Reranker-4B} reranker for fair comparison.

\paragraph{Dense X Retrieval~\cite{chen2024densex}} decomposes documents into fine-grained propositions and uses these propositions as retrieval units. This improves precision by making each unit more atomic and self-contained. However, proposition-level retrieval can weaken broader contextual continuity, since related facts are retrieved independently rather than as coherent multi-sentence evidence.

\paragraph{Meta-Chunking~\cite{zhao2024metachunking}} uses LLM-based signals to identify semantically meaningful chunk boundaries instead of relying on fixed-size segmentation. We evaluate both variants: \textit{Meta-Chunking-PPL}, which uses perplexity changes to detect boundaries, and \textit{Meta-Chunking-MSP}, which uses margin-sampling-based boundary decisions. These methods improve chunk coherence, but once chunks are formed, cross-chunk semantic dependencies are not explicitly modeled.

\paragraph{MoC~\cite{zhao-etal-2025-moc}} improves over single-strategy chunking by dynamically routing text to different chunking strategies or granularity choices. It is a strong adaptive chunking baseline because it can better match the segmentation strategy to the local document structure. However, MoC still primarily operates at the chunk-construction stage and does not build a retrieval-time hierarchy over sentence-level evidence.

\paragraph{RAPTOR~\cite{sarthi2024raptor}} recursively clusters chunks and summarizes each cluster to build a hierarchical tree. Retrieval can then operate over both lower-level chunks and higher-level summaries. This provides a natural multi-granularity baseline, but its hierarchy is based on embedding-space clustering and LLM-generated summaries, which can introduce information loss and additional indexing cost.

\paragraph{LightRAG~\cite{guo2025lightrag}} augments retrieval with graph-structured knowledge and combines local and global retrieval signals. It is designed to improve retrieval over connected evidence by exploiting entity and relation structure. We include it as a structured retrieval baseline, especially for settings where graph-style evidence organization can improve coverage.

\paragraph{PropRAG~\cite{wang2025proprag}} represents documents using propositions and performs beam-style traversal over proposition paths. It is closely related to our use of beam search, but differs in its underlying structure: PropRAG searches over a proposition graph, whereas \name{} searches over an attention-guided sentence hierarchy. PropRAG is therefore a strong baseline for testing whether hierarchical sentence-level structure provides benefits beyond proposition-path retrieval.

\paragraph{SAKI-RAG~\cite{tao2025sakirag}} uses a Sentence-Level Large Language Model (SLLM) to estimate inter-sentence attention and expand retrieved chunks with related sentences. It is the closest baseline to \name{} because it also uses sentence-level attention signals. However, SAKI-RAG performs pairwise expansion and relies on LLM filtering during retrieval, while \name{} converts learned attention into a global tree and retrieves across multiple granularities without inference-time LLM filtering.

\paragraph{GraphRAG~\cite{edge2024graphrag}} constructs an entity-relation graph from the corpus and uses graph structure to support retrieval and generation. It is effective for queries that align well with entity-centric evidence, but it depends on reliable entity extraction and relation construction. We include GraphRAG in the end-to-end comparison because it is a system-level RAG method with substantial LLM-based preprocessing and reasoning.

\paragraph{ReflectiveRAG~\cite{verma-etal-2026-reflectiverag}} introduces adaptive retrieval and generation through self-reflection. Instead of using a fixed retrieval budget, it evaluates whether retrieved evidence is sufficient and can reformulate or expand retrieval when needed. This makes it a strong final-performance baseline, but its main contribution lies in adaptive evidence use and generation rather than efficient retrieval structure.

\paragraph{PageIndex~\cite{zhang2025pageindex}} replaces conventional vector retrieval with a reasoning-based hierarchical page index. An LLM navigates the index and selects evidence through multi-step reasoning, which can improve final answer quality. However, because it performs LLM-heavy online search and evidence construction, we include it only in end-to-end performance comparisons rather than as an efficient retrieval baseline.

\paragraph{REFRAG~\cite{lin2025refrag}} focuses on generation-side efficiency for RAG by exploiting sparsity in retrieved contexts. It compresses retrieved chunks and selectively expands them during decoding, reducing the effective context processed by the generator. Since its main contribution is decoding-time optimization rather than retrieval or indexing, we include it in the final-performance and efficiency comparison.
\section{Retrieval Performance at Different Cutoffs}
\label{app:detailed_results}

Tables~\ref{tab:results_at1}, \ref{tab:results_at3}, and \ref{tab:results_at5} report retrieval performance at cutoffs $k{=}1$, $k{=}3$, and $k{=}5$, respectively, across all six benchmarks. As expected, both precision and recall increase monotonically with $k$ for all methods, since retrieving more documents provides greater coverage of relevant passages. The relative ordering of methods remains consistent across all cutoffs: \name{} outperforms all non-oracle baselines at every depth. This consistency demonstrates that the gains from topic-guided retrieval are not specific to any particular cutoff, but reflect a robust improvement in retrieval quality across the full range of evaluation settings reported here.

\begin{table*}[t]
    \centering
    \setlength{\tabcolsep}{4.5pt}
    \renewcommand{\arraystretch}{1.1}
    
    \resizebox{\textwidth}{!}{%
    \begin{tabular}{lcccccccccccc}
    \toprule
    \multirow{2}{*}{\textbf{Method}}
    & \multicolumn{3}{c}{\textbf{LegalBench-RAG}} & \multicolumn{3}{c}{\textbf{MSMARCO}} & \multicolumn{3}{c}{\textbf{SCI-DOCS}} & \multicolumn{3}{c}{\textbf{Dragonball}} \\
    \cmidrule(lr){2-4}\cmidrule(lr){5-7}\cmidrule(lr){8-10}\cmidrule(lr){11-13}
      & \textit{IE$\uparrow$} & \textit{Prec.$\uparrow$} & \textit{Rec.$\uparrow$}  & \textit{IE$\uparrow$} & \textit{Prec.$\uparrow$} & \textit{Rec.$\uparrow$}  & \textit{IE$\uparrow$} & \textit{Prec.$\uparrow$} & \textit{Rec.$\uparrow$}  & \textit{IE$\uparrow$} & \textit{Prec.$\uparrow$} & \textit{Rec.$\uparrow$} \\
    \midrule
    DenseXRetrieval
      & 1.93 & 32.74 & 5.88
      & 19.14 & 58.68 & 32.61
      & 70.89 & 86.13 & 82.31
      & 0.01 & 1.64 & 1.12 \\
    Meta-Chunking-PPL
      & 2.72 & 35.41 & 7.69
      & 20.94 & 59.97 & 34.92
      & 0.36 & 16.73 & 2.15
      & 6.72 & 41.62 & 16.14 \\
    Meta-Chunking-MSP
      & 2.94 & 36.26 & 8.12
      & 21.57 & 60.36 & 35.73
      & 71.91 & 86.42 & 83.21
      & 4.21 & 39.08 & 10.77 \\
    RAPTOR
      & 3.10 & 36.83 & 8.41
      & 22.94 & 61.81 & 37.12
      & 80.94 & 91.03 & 88.92
      & 4.08 & 38.92 & 10.49 \\
    LightRAG
      & 4.29 & 40.97 & 10.47
      & 26.57 & 64.42 & 41.25
      & 80.42 & 90.84 & 88.53
      & 11.60 & 51.16 & 22.67 \\
    PropRAG
      & 4.76 & 42.28 & 11.25
      & 28.13 & 65.83 & 42.73
      & 83.63 & 92.36 & 90.55
      & 14.37 & 55.24 & 26.01 \\
    MoC
      & \underline{5.10} & 43.11 & \underline{11.84}
      & 28.68 & 66.26 & 43.29
      & \underline{84.39} & 92.73 & \underline{91.01}
      & 16.09 & 57.47 & \underline{27.99} \\
    SAKI-RAG
      & 4.98 & \underline{43.27} & 11.52
      & \underline{30.21} & \underline{67.58} & \underline{44.70}
      & 83.46 & \underline{95.43} & 87.46
      & \underline{19.62} & \underline{75.61} & 25.95 \\
    \midrule
    \rowcolor{blue!8}
    \name{}
      & \textbf{8.45} & \textbf{50.48} & \textbf{16.74}
      & \textbf{37.21} & \textbf{72.21} & \textbf{51.53}
      & \textbf{93.91} & \textbf{98.04} & \textbf{95.79}
      & \textbf{29.68} & \textbf{82.34} & \textbf{36.05} \\
    \bottomrule
    \end{tabular}
    }% end resizebox
    \caption{Retrieval performance at depth $k{=}1$ across four benchmarks (IE\,$\uparrow$, Precision\,$\uparrow$, Recall\,$\uparrow$). \textbf{Bold} = best; \underline{underline} = second-best. \name{} rows are shaded.}
    \label{tab:results_at1}
\end{table*}

\begin{table*}[t]
    \centering
    \setlength{\tabcolsep}{4.5pt}
    \renewcommand{\arraystretch}{1.1}
    
    \resizebox{\textwidth}{!}{%
    \begin{tabular}{lcccccccccccc}
    \toprule
    \multirow{2}{*}{\textbf{Method}}
    & \multicolumn{3}{c}{\textbf{LegalBench-RAG}} & \multicolumn{3}{c}{\textbf{MSMARCO}} & \multicolumn{3}{c}{\textbf{SCI-DOCS}} & \multicolumn{3}{c}{\textbf{Dragonball}} \\
    \cmidrule(lr){2-4}\cmidrule(lr){5-7}\cmidrule(lr){8-10}\cmidrule(lr){11-13}
      & \textit{IE$\uparrow$} & \textit{Prec.$\uparrow$} & \textit{Rec.$\uparrow$}  & \textit{IE$\uparrow$} & \textit{Prec.$\uparrow$} & \textit{Rec.$\uparrow$}  & \textit{IE$\uparrow$} & \textit{Prec.$\uparrow$} & \textit{Rec.$\uparrow$}  & \textit{IE$\uparrow$} & \textit{Prec.$\uparrow$} & \textit{Rec.$\uparrow$} \\
    \midrule
    DenseXRetrieval
      & 2.84 & 34.99 & 8.13
      & 21.96 & 61.68 & 35.61
      & 73.44 & 87.63 & 83.81
      & 0.01 & 5.01 & 0.14 \\
    Meta-Chunking-PPL
      & 3.74 & 37.66 & 9.94
      & 23.88 & 62.97 & 37.92
      & 0.67 & 18.23 & 3.65
      & 8.07 & 43.87 & 18.39 \\
    Meta-Chunking-MSP
      & 3.99 & 38.51 & 10.37
      & 24.54 & 63.36 & 38.73
      & 74.48 & 87.92 & 84.71
      & 5.38 & 41.33 & 13.02 \\
    RAPTOR
      & 4.17 & 39.08 & 10.66
      & 26.00 & 64.81 & 40.12
      & 83.67 & 92.53 & 90.42
      & 5.25 & 41.17 & 12.74 \\
    LightRAG
      & 5.50 & 43.22 & 12.72
      & 29.83 & 67.42 & 44.25
      & 83.13 & 92.34 & 90.03
      & 13.31 & 53.41 & 24.92 \\
    PropRAG
      & 6.01 & 44.53 & 13.50
      & 31.48 & 68.83 & 45.73
      & 86.40 & 93.86 & 92.05
      & 16.25 & 57.49 & 28.26 \\
    MoC
      & \underline{6.39} & 45.36 & \underline{14.09}
      & 32.06 & 69.26 & 46.29
      & \underline{87.17} & 94.23 & \underline{92.51}
      & 18.06 & 59.72 & \underline{30.24} \\
    SAKI-RAG
      & 6.27 & \underline{45.52} & 13.77
      & \underline{33.67} & \underline{70.58} & \underline{47.70}
      & 86.23 & \underline{96.93} & 88.96
      & \underline{21.96} & \underline{77.86} & 28.20 \\
    \midrule
    \rowcolor{blue!8}
    \name{}
      & \textbf{10.01} & \textbf{52.73} & \textbf{18.99}
      & \textbf{41.01} & \textbf{75.21} & \textbf{54.53}
      & \textbf{95.92} & \textbf{98.69} & \textbf{97.19}
      & \textbf{32.40} & \textbf{84.59} & \textbf{38.30} \\
    \bottomrule
    \end{tabular}
    }% end resizebox
    \caption{Retrieval performance at depth $k{=}3$ across four benchmarks (IE\,$\uparrow$, Precision\,$\uparrow$, Recall\,$\uparrow$). \textbf{Bold} = best; \underline{underline} = second-best. \name{} rows are shaded.}
    \label{tab:results_at3}
\end{table*}

\begin{table*}[t]
    \centering
    \setlength{\tabcolsep}{4.5pt}
    \renewcommand{\arraystretch}{1.1}
    
    \resizebox{\textwidth}{!}{%
    \begin{tabular}{lcccccccccccc}
    \toprule
    \multirow{2}{*}{\textbf{Method}}
    & \multicolumn{3}{c}{\textbf{LegalBench-RAG}} & \multicolumn{3}{c}{\textbf{MSMARCO}} & \multicolumn{3}{c}{\textbf{SCI-DOCS}} & \multicolumn{3}{c}{\textbf{Dragonball}} \\
    \cmidrule(lr){2-4}\cmidrule(lr){5-7}\cmidrule(lr){8-10}\cmidrule(lr){11-13}
      & \textit{IE$\uparrow$} & \textit{Prec.$\uparrow$} & \textit{Rec.$\uparrow$}  & \textit{IE$\uparrow$} & \textit{Prec.$\uparrow$} & \textit{Rec.$\uparrow$}  & \textit{IE$\uparrow$} & \textit{Prec.$\uparrow$} & \textit{Rec.$\uparrow$}  & \textit{IE$\uparrow$} & \textit{Prec.$\uparrow$} & \textit{Rec.$\uparrow$} \\
    \midrule
    DenseXRetrieval
      & 4.99 & 39.49 & 12.63
      & 28.16 & 67.68 & 41.61
      & 78.68 & 90.63 & 86.81
      & 0.04 & 9.51 & 0.40 \\
    Meta-Chunking-PPL
      & 6.09 & 42.16 & 14.44
      & 30.29 & 68.97 & 43.92
      & 1.41 & 21.23 & 6.65
      & 11.07 & 48.37 & 22.89 \\
    Meta-Chunking-MSP
      & 6.40 & 43.01 & 14.87
      & 31.02 & 69.36 & 44.73
      & 79.75 & 90.92 & 87.71
      & 8.03 & 45.83 & 17.52 \\
    RAPTOR
      & 6.61 & 43.58 & 15.16
      & 32.66 & 70.81 & 46.12
      & 89.24 & 95.53 & 93.42
      & 7.87 & 45.67 & 17.24 \\
    LightRAG
      & 8.22 & 47.72 & 17.22
      & 36.89 & 73.42 & 50.25
      & 88.69 & 95.34 & 93.03
      & 17.04 & 57.91 & 29.42 \\
    PropRAG
      & 8.83 & 49.03 & 18.00
      & 38.71 & 74.83 & 51.73
      & 92.07 & 96.86 & 95.05
      & 20.31 & 61.99 & 32.76 \\
    MoC
      & \underline{9.27} & 49.86 & \underline{18.59}
      & 39.35 & 75.26 & 52.29
      & \underline{92.86} & 97.23 & \underline{95.51}
      & 22.31 & 64.22 & \underline{34.74} \\
    SAKI-RAG
      & 9.14 & \underline{50.02} & 18.27
      & \underline{41.12} & \underline{76.58} & \underline{53.70}
      & 91.90 & \underline{99.93} & 91.96
      & \underline{26.93} & \underline{82.36} & 32.70 \\
    \midrule
    \rowcolor{blue!8}
    \name{}
      & \textbf{13.44} & \textbf{57.23} & \textbf{23.49}
      & \textbf{49.16} & \textbf{81.21} & \textbf{60.53}
      & \textbf{100.00} & \textbf{100.00} & \textbf{100.00}
      & \textbf{38.13} & \textbf{89.09} & \textbf{42.80} \\
    \bottomrule
    \end{tabular}
    }% end resizebox
    \caption{Retrieval performance at depth $k{=}5$ across four benchmarks (IE\,$\uparrow$, Precision\,$\uparrow$, Recall\,$\uparrow$). \textbf{Bold} = best; \underline{underline} = second-best. \name{} rows are shaded.}
    \label{tab:results_at5}
\end{table*}

\section{Qualitative Analysis: Recovering Multi-Sentence Legal Evidence}
\label{app:qualitative_legal}

Table~\ref{tab:qualitative_tree_node} presents a qualitative example that illustrates why retrieving \textbf{internal tree nodes} is useful. The query asks whether a software services agreement limits the provider's liability and what exceptions apply. A complete answer requires more than a single sentence or a pair of related sentences: the model must recover the excluded damages, the aggregate liability cap, the scope of the limitation across legal theories, and the carve-outs for indemnification, confidentiality breaches, gross negligence, and willful misconduct. These pieces form a coherent clause-level unit, but they are distributed across four sentences. MoC retrieves a locally coherent chunk containing the liability cap, but this evidence is too narrow to answer the full query. SAKI-RAG improves over local chunking by linking the damage-exclusion sentence with the liability-cap sentence; however, its pairwise expansion still misses the later exception sentence, which is essential for a legally complete answer. In contrast, \name{} retrieves the internal node $v_{1:4}$, which groups all four relevant sentences into a single clause-level unit. This allows the generator to answer both parts of the query: the agreement limits liability through damage exclusions and a monetary cap, but the limitation does not apply to the specified carve-outs.

\begin{table*}[t]
\centering
\small
\resizebox{\textwidth}{!}{%
\begin{tabular}{p{0.17\textwidth} p{0.43\textwidth} p{0.32\textwidth}}
\toprule
\textbf{Method / Unit} & \textbf{Retrieved Evidence} & \textbf{Analysis} \\
\midrule
\rowcolor{gray!12}
\multicolumn{3}{c}{
\textit{\textbf{Query:} Does the agreement limit the provider's liability, and what exceptions or exclusions apply?}
} \\
\midrule

MoC 
& 
\textbf{Retrieved chunk: liability cap.}
\textit{Provider's aggregate liability under this Agreement shall not exceed the fees paid by Client during the twelve (12) months preceding the event giving rise to the claim.}
& 
MoC identifies a locally coherent chunk around the monetary cap, but the retrieved unit is too narrow for the query. It answers \textit{how much} liability is capped, but misses the excluded damages, the scope across legal theories, and the exceptions. \\

SAKI-RAG 
& 
\textbf{Retrieved pairwise expansion: damage exclusion + liability cap.}
\begin{enumerate}
    \item \textit{In no event shall Provider be liable for any indirect, incidental, special, consequential, exemplary, or punitive damages.}
    \item \textit{Provider's aggregate liability shall not exceed the fees paid during the prior twelve months.}
\end{enumerate}
& 
SAKI-RAG improves over a single chunk by linking two related sentences. However, the evidence remains pairwise, so it captures the main limitation but misses the later sentence listing exceptions such as indemnification, confidentiality breach, gross negligence, and willful misconduct. \\

\midrule
Leaf $s_1$
&
\textit{In no event shall Provider be liable for any indirect, incidental, special, consequential, exemplary, or punitive damages arising out of or relating to this Agreement.}
&
Identifies excluded damages, but does not provide the monetary cap or exceptions. \\

Leaf $s_2$
&
\textit{Provider's aggregate liability under this Agreement shall not exceed the fees paid by Client during the twelve (12) months preceding the event giving rise to the claim.}
&
Provides the liability cap, but not the scope or carve-outs. \\

Leaf $s_3$
&
\textit{The foregoing limitation shall apply regardless of the form of action, whether in contract, tort, strict liability, or otherwise.}
&
Clarifies that the limitation applies across legal theories. \\

Leaf $s_4$
&
\textit{The limitations in this Section shall not apply to Provider's indemnification obligations, breach of confidentiality, gross negligence, or willful misconduct.}
&
Provides the exceptions required for a complete legal answer. \\

\midrule
\rowcolor{blue!8}
\textbf{\name{} internal node $v_{1:4}$}
&
\textbf{Retrieved clause-level node containing $s_1$--$s_4$:}
\begin{enumerate}
    \item excluded damages,
    \item aggregate liability cap,
    \item scope across legal theories,
    \item exceptions and carve-outs.
\end{enumerate}
&
\textbf{\name{} retrieves the middle node containing more than two sentences.} This gives the generator the full limitation-of-liability clause, enabling a complete answer that includes both the limitation and the exceptions. \\
\bottomrule
\end{tabular}%
}
\caption{
Qualitative comparison on a limitation-of-liability query. MoC retrieves a locally coherent but incomplete chunk, and SAKI-RAG retrieves a related sentence pair that captures the main limitation but misses the exception sentence. \name{} retrieves an internal clause-level node containing four sentences, allowing the answer to include excluded damages, the liability cap, legal-theory scope, and carve-outs.
}
\label{tab:qualitative_tree_node}
\end{table*}

\end{document}